\documentclass[conference]{IEEEtran}
\IEEEoverridecommandlockouts
\usepackage{amsmath}
\usepackage{graphicx}
\usepackage{tikz}
\usetikzlibrary{arrows}
\makeatletter

\usepackage{mathtools,dsfont}
\usepackage{psfrag,bbm,graphicx}
\usepackage{comment,balance}
\usepackage{epstopdf}
\usepackage{epsfig,subfig}
\usepackage{multicol}
\usepackage{multirow, cite}
\usepackage{amsfonts, color}
\usepackage{array}
\usepackage{amssymb}
\usepackage[linesnumbered,ruled,vlined]{algorithm2e}
\usepackage{tikz}
\usetikzlibrary{automata, positioning}
\usepackage{etoolbox}
\usepackage{geometry}
\usepackage{hyperref}
\usepackage{amsthm}

\patchcmd{\thebibliography}{\section*{\refname}}{}{}{}    

\newtheorem{lemma}{Lemma}
\newtheorem*{lemma*}{Lemma} 
\newtheorem{remark}{Remark}

\newtheorem{theorem}{Theorem}

\newcommand{\remove}[1]{{}}

\newcommand\numberthis{\addtocounter{equation}{1}\tag{\theequation}}

\newcommand\blfootnote[1]{%
	\begingroup
	\renewcommand\thefootnote{}\footnote{#1}%
	\addtocounter{footnote}{-1}%
	\endgroup
}

\def\E{{\mathbb{E}}}
\def\Prob{{\mathbb{P}}}
\def\Indc{{\mathbbm{1}}}

\def \OO {\mathrm{O}}

\makeatletter
\def\endthebibliography{%
	\def\@noitemerr{\@latex@warning{Empty `thebibliography' environment}}%
	\endlist
}
\makeatother

\newfont{\mycrnotice}{ptmr8t at 7pt}
\newfont{\myconfname}{ptmri8t at 7pt}

\title{Cascading Bandits With Feedback}
\author{\IEEEauthorblockN{R Sri Prakash}
\IEEEauthorblockA{IIITDM Kancheepuram} 
sriprakash@iiitdm.ac.in
\and 
\IEEEauthorblockN{ Nikhil Karamchandani}
\IEEEauthorblockA{IIT Bombay} 
nikhilk@ee.iitb.ac.in
\and
\IEEEauthorblockN{ Sharayu Moharir} 
\IEEEauthorblockA{IIT Bombay}
sharayum@ee.iitb.ac.in
}

\allowdisplaybreaks

\def\BibTeX{{\rm B\kern-.05em{\sc i\kern-.025em b}\kern-.08em
		T\kern-.1667em\lower.7ex\hbox{E}\kern-.125emX}}

\begin{document}
\maketitle        
\begin{abstract}
Motivated by the challenges of edge inference, we study a variant of the cascade bandit model in which each arm corresponds to an inference model with an associated accuracy and error probability.  We analyse four decision-making policies—Explore-then-Commit, Action Elimination, Lower Confidence Bound (LCB), and Thompson Sampling—and provide sharp theoretical regret guarantees for each. Unlike in classical bandit settings, Explore-then-Commit and Action Elimination incur suboptimal regret because they commit to a fixed ordering after the exploration phase, limiting their ability to adapt. In contrast, LCB and Thompson Sampling continuously update their decisions based on observed feedback, achieving constant $\mathcal{O}(1)$ regret. Simulations corroborate these theoretical findings, highlighting the crucial role of adaptivity for efficient edge inference under uncertainty. \blfootnote{Nikhil Karamchandani's work was supported by a SERB MATRICS grant.}
\end{abstract}

\begin{IEEEkeywords}
cascade bandits
\end{IEEEkeywords}

\section{Introduction}
The increasing penetration of Artificial Intelligence (AI) and the Internet of Things (IoT) across diverse application domains has led to a growing demand for Mobile Edge Computing (MEC). To achieve faster inference from models trained for specific tasks, it is essential to host them at the edge, thereby reducing latency and improving responsiveness.  In this work, we adopt the approach of using a collection of simpler, task-specific models organized in a cascade to perform inference at the edge. A user submits an inference request to the edge, thereby alleviating the computational burden on the User Equipment (UE).

\begin{figure}[h]
    \centering
    \includegraphics[width=0.8\linewidth]{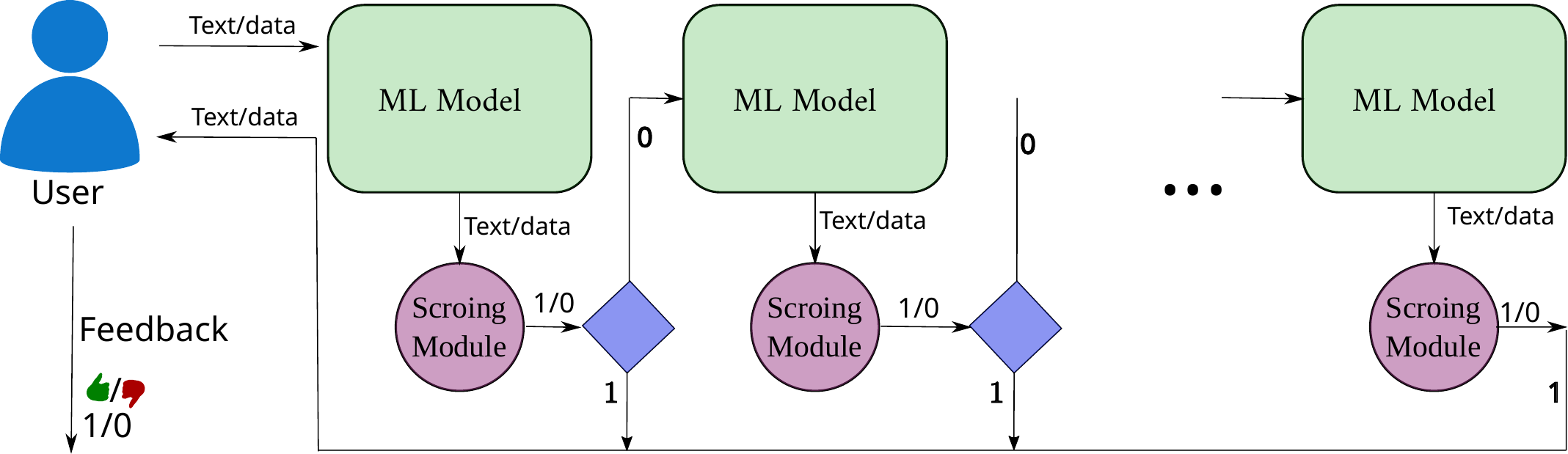}
    \caption{Cascade ML models with scoring modules}
    \label{fig:network}
\end{figure}

Our system consists of multiple cascaded machine learning (ML) models, each paired with a dedicated scoring module, as shown in \figurename~\ref{fig:network}, following a design similar to \cite{chen2023frugalgpt}. For a given query, each ML model generates an output, which is then evaluated by its corresponding scoring module. The scoring module maps the model’s output to a binary decision: a value of one indicates that the response is capable of satisfying the user’s query, while a value of zero indicates otherwise. If the scoring module outputs one, the response from that ML model is forwarded to the user, who subsequently provides binary feedback indicating whether the response is satisfactory (one) or unsatisfactory (zero). Conversely, if the scoring module outputs zero, the query is passed to the next ML model in the cascade. This process continues until a scoring module outputs one or until all scoring modules return zero. In the latter case, the cascade model is unable to satisfy the user’s query.

A critical consideration in our model is that a scoring module's output of 1 does not guarantee a satisfactory user experience, as the user may still provide feedback of 0. We account for this by modeling a stochastic error probability for each scoring module. This error is only detectable when a scoring module returns a 1, but the user's subsequent feedback is 0. The reward obtained in this setting is defined as 1 if the user feedback is satisfactory, and 0 if the feedback is unsatisfactory or if the system fails to serve the request. The central objective of this work is to determine the optimal ordering of the ML models within the cascade so as to maximize the total reward or, equivalently, to minimize the regret, defined as the difference between the reward of the policy under consideration and that of the static optimal policy.

We model our edge inference setup as a cascade of ML models, each paired with a scoring module as a cascade bandit problem with feedback. Each model--scorer pair is treated as a bandit arm, and the system output corresponds to the first model in the cascade whose scorer outputs one. This setting is closely related to the cascade bandit framework \cite{kveton2015cascading,zong2016cascading}, where a list of items is presented to a user and a reward is obtained if an item is clicked. In our case, the agent selects the first model with a positive score, presents its response, and receives binary user feedback: one if satisfactory, zero otherwise. If all scorers output zero, the request is unserved, analogous to the ``no-click'' outcome in cascade bandits. Unlike \cite{zhong2020best}, which studies best-arm identification, we focus on regret minimisation; and unlike \cite{wang2023cascading}, our feedback is immediate rather than randomly delayed.  

Our model differs from the conventional cascade bandit setting in that the reward depends directly on the order of models in the cascade, unlike the standard setup, where order does not affect slot reward. It is also related to cost-aware cascading bandits \cite{gan2020cost,tran2012knapsack,burnetas2012adaptive,ding2013multi}, where ordering impacts the total reward due to examination costs. The closest prior work is \cite{cheng2022cascading}, where an agent presents a list of items and optimises the order to maximise reward. While our LCB algorithm differs from their UCB variant, we adapt their theoretical techniques and further leverage results from \cite{audibert2010best,wang18a,agrawal2017near,zhong2021thompson} to establish sharp regret guarantees.

The key contributions of this work are as follows. We show that any policy which becomes static after a certain number of time slots—such as explore-and-commit algorithms—will incur a regret of order $\Omega(\log T)$. We then analyze the performance of the Lower Confidence Bound (LCB) algorithm and the Thompson Sampling algorithm in our cascade bandit setting, and demonstrate that both achieve $\mathrm{O}(1)$ regret. Furthermore, we validate these theoretical results through extensive simulations, showing strong consistency between the theoretical guarantees and empirical performance.



\section{Problem Setup}
We consider a $K$-arm cascade bandit problem, where the set of arms is denoted by $[K]=\{1,2,\dots, K\}$. Each arm produces a binary output independently of the others. Let $X_i(t)$ denote the output of arm $i$ at time slot $t$ (representing the scoring module’s decision), where $\{X_i(t)\}_{t\geq 1}$ are i.i.d.\ Bernoulli random variables with mean $\mu_i = \E[X_i(t)]$.  

At each time slot $t$, the learner selects an ordering of the arms, denoted by $\mathcal{L}_t = (l_1^{(t)}, l_2^{(t)}, \dots, l_K^{(t)})$, where $l_j^{(t)}$ denotes the arm placed in position $j$ of the cascade. The cascade is traversed sequentially in this order until the first arm that outputs one is encountered. Formally, the selected arm index at time $t$ is $I_t = \min\{ j \in [K] : X_{l_j^{(t)}}(t)=1\}$, with the convention that $I_t = \infty$ if all arms output zero.  

The response of the selected arm $l_{I_t}^{(t)}$ is shown to the user, who provides feedback $Y_{l_{I_t}^{(t)}}(t)\in\{0,1\}$ indicating whether the displayed result was relevant. The user feedback is treated as the reward in slot $t$. If $I_t=\infty$, no arm is displayed and the reward is zero. An arm $i$ is thus observed only if $X_i(t)=1$ and all arms preceding it in the ordering $\mathcal{L}_t$ output zero.  

An error is said to occur if $X_i(t)=1$ and $Y_i(t)=0$. These errors are stochastic, with error probability defined as $p_i = \Prob(Y_i(t)=0 \mid X_i(t)=1), \quad \forall i \in [K]$. Without loss of generality, we assume the arms are indexed such that $p_1 < p_2 < \cdots < p_K$. For notational convenience, we also define the gaps $\Delta_i = p_i - p_{i-1}$ for $2 \leq i \leq K$.  

The goal is to maximize the cumulative reward, or equivalently, minimize the cumulative regret, by dynamically adapting the ordering of the arms in the cascade. Let $\mathcal{L}^*=(l_1^*, l_2^*, \dots, l_K^*)$ denote the optimal ordering of the arms, which sorts arms in increasing order of their error probabilities. Furthermore, let $l_t^{-1}(i)$ denote the position of arm $i$ in the ordering $\mathcal{L}_t$. The regret of a policy then quantifies the expected reward loss incurred relative to always playing $\mathcal{L}^*$.

We define the expected reward of a cascade model when the arms are ordered according to the ordering $\mathcal{L}=(l_1,l_2,\dots,l_K)$ as
\begin{align*}
	r_\mathcal{L}=\sum_{i=1}^{K}(1-p_{l_i})\mu_{l_i}\prod_{j=1}^{i-1}(1-\mu_{l_j}).
\end{align*}

This expression reflects that the $i$-th arm in the ordering contributes to the reward only if all preceding arms $1,\dots,i-1$ fail to produce an output $1$, which occurs with probability $\prod_{j=1}^{i-1}(1-\mu_{l_j})$. Given that arm $i$ is shown, it produces a satisfactory feedback with probability $(1-p_{l_i})$, and the event of the arm being triggered occurs with probability $\mu_{l_i}$. Hence, the summation accounts for the expected contribution of each arm to the overall reward under ordering $\mathcal{L}$.

We define the suboptimality gap of an ordering $\mathcal{L}$ as $\tilde{\Delta}_\mathcal{L}= r_{\mathcal{L}^*}-r_{\mathcal{L}}$, which quantifies the loss in expected reward when using $\mathcal{L}$ instead of the optimal ordering $\mathcal{L}^*$. For each arm $i$, let $\mathcal{M}_i$ be the set of all orderings starting with $i$, and define $\tilde{\Delta}_i=\max_{\mathcal{L}\in \mathcal{M}_i}\tilde{\Delta}_{\mathcal{L}}$ as the maximum suboptimality incurred by choosing $i$ as the first arm. The largest such value across arms, $\tilde{\Delta}_{max}=\max_i\tilde{\Delta}_i$, represents the worst-case loss from starting with a poor arm, while the smallest nonzero gap, $\tilde{\Delta}_{min}=\min_{\mathcal{L}\ne \mathcal{L^*}}\tilde{\Delta}_\mathcal{L}$, captures how hard it is to distinguish the optimal ordering from its closest suboptimal alternative.

Regret is defined as the difference in expected reward between the optimal policy and the policy under consideration. Let $\mathcal{R}^{\pi}(T)$ represent the expected regret of policy $\pi$ until time $T$, and let $\mathcal{R}_t^{\pi}$ denote the expected regret incurred in slot $t$:


\begin{align*}
	\mathcal{R}_t^{\pi}=&r_{\mathcal{L}^*} -  r_{\mathcal{L}_t}\\
	\mathcal{R}^{\mathcal{\pi}}(T) =& \sum_{t=1}^{T} \mathcal{R}_t^{\pi}
\end{align*}

The objective is to select the order of arms $\mathcal{L}_t$ in each round to minimize the cumulative regret. This essentially means we want to find a policy that, over time, consistently chooses an arm ordering that is as close as possible to the optimal one.


\section{Results} \label{sec:results}
In this section, we will first define the optimal static policy for our problem and then introduce several online algorithms designed to minimise regret. These algorithms—Explore and Commit (EC), Action Elimination (AE), Lower Confidence Bound (LCB), and Thompson Sampling (TS) — each come with specific performance guarantees.
\subsection{Static Optimal Policy}
\begin{theorem}\label{thm:opt_policy}
The optimal static policy will order the arms in increasing order of their error probabilities $(p_i)$
\end{theorem}
The proof of Theorem \ref{thm:opt_policy} is provided in the Appendix \cite{techdoc}.
\begin{remark}
    The optimal ordering of arms is independent of their means $(\mu_i)$, which are used for sampling. Instead, the ordering relies exclusively on the probability of error $(p_i)$, meaning arms with a lower error rate are given priority and positioned first
\end{remark}

\subsection{Explore and Commit}
The Explore and Commit (EC) algorithm, formally described in Algorithm~\ref{Alg:Exp_cmmt}, is a two-phase strategy designed to balance exploration with exploitation in the cascade bandit setting. It begins with an exploration phase, where each arm is pulled an equal number of times in order to estimate its probability of error. The number of pulls per arm is defined as $N = \max_i n_i$, where $n_i = \lceil \tfrac{16 \log T}{\Delta_i^2 \mu_i} \rceil$, and the total exploration period lasts for $T_s^{EC} = NK$ slots. During this phase, the algorithm cycles through the arms in a rotating order, ensuring that each arm appears in every position of the cascade. At each time slot, the cascade presents the output of the first triggered arm to the user, whose feedback is then used to update the empirical error probability of the selected arm and to compute lower confidence bounds on these estimates. Once the exploration phase is complete, the algorithm enters the commit phase, where the arms are permanently ordered in ascending order of their estimated error probabilities. This committed ordering is then used for the remainder of the horizon. 
\begin{algorithm}
	\SetAlgoLined
	\KwIn{$\Delta_i$, $T$, $\mu_i$}
	\KwOut{Ordered list $\mathcal{L}_t$}
	Initialize: $\mathcal{L}_1=$ random order of arms, $t=1$\\
    \quad $N=\max_i\lceil \tfrac{16 \log T}{\Delta_i^2 \mu_i} \rceil$\\
		\textbf{Explore:}\\
			\For{$t\le NK$}{
				Order arms according to $\mathcal{L}_{t}$\\
				The cascade model shows the result of arm $I_t$ to the user \\
				Feedback $Y_{I_t}(t)$ is observed\\
				\For{$i\in[K]$}{
					$S_{i}(t)=S_{i}(t-1)+\mathbbm{1}_{\{i=I_t\}}$\\
                    $\hat{p}_{i}(t)=\left\{\sum_{n=1}^t \mathbbm{1}_{\{i=I_n\}}(1-Y_{i}(t))\right\}/S_i(t)$\\
					$L_i(t)=\hat{p}_i(t)-\sqrt{\frac{2 \log T}{S_i(t)}}$
				}	
				$\mathcal{L}_{t+1}=(l_2^{(t)},l_3^{(t)},\cdots,l^{(t)}_{K},l_1^{(t)})$\\
				$t++$}
		
		\textbf{Commit:}\\
		$\mathcal{L}'$= List arms in ascending order of $L_{i}(t-1)$\\
		\For{$t<T$}{$\mathcal{L}_t=\mathcal{L}'$\\
			$t++$
		}
	\caption{Explore and Commit}
	\label{Alg:Exp_cmmt}
\end{algorithm}

\begin{lemma} \label{lem:EC_reg}
	The probability that Algorithm~\ref{Alg:Exp_cmmt} selects a suboptimal ordering in the commit phase is bounded by
	\[
		\Prob(\mathcal{L}_t \ne \mathcal{L}^*) \;\le\; \frac{K}{T^{2}} + \frac{K^2}{T^4}.
	\]
\end{lemma}

\noindent The proof of Lemma~\ref{lem:EC_reg} is provided in the Appendix\cite{techdoc}.


\begin{theorem}
	The regret obtained by Algorithm~\ref{Alg:Exp_cmmt} satisfies
	\[
		\mathcal{R}^{EC}(T) = \OO(\log T).
	\]
\end{theorem}

\begin{proof}
The total regret can be decomposed into contributions from the exploration phase and the commit phase:
\begin{align*}
	\mathcal{R}^{EC}(T) 
	&= \sum_{t=1}^{T_s^{EC}} \E[\mathcal{R}_t] 
	+ \sum_{t=T_s^{EC}+1}^T \Prob(\mathcal{L}_t \ne \mathcal{L}^*) \, \tilde{\Delta}_{\mathcal{L}_t}. 
\end{align*}
Using Lemma~\ref{lem:EC_reg}, we obtain
\begin{align*}
	\mathcal{R}^{EC}(T) 
	&\leq NK \, \tilde{\Delta}_{\max} 
	+ T \tilde{\Delta}_{\max} \left( \frac{K}{T^{2}} + \frac{K^2}{T^4} \right).
\end{align*}
Since $N = \OO(\log T)$, the first term scales as $\OO(\log T)$, and the second term is asymptotically negligible.
Hence,
	$$\mathcal{R}^{EC}(T) = \OO(\log T).$$
\end{proof}


\begin{theorem}\label{thm:EC_lb}
	The regret of Algorithm~\ref{Alg:Exp_cmmt} is lower bounded as	
		$$\mathcal{R}^{EC}(T) = \Omega(\log T).$$	
\end{theorem}

The proof of Theorem \ref{thm:EC_lb} is provided in Appendix \cite{techdoc}.


However, a key limitation of EC is that the required number of pulls $N$ depends on the gap parameters $\Delta_i$, which are generally unknown in practice, making the algorithm difficult to implement in real-world settings. This naturally motivates the use of adaptive strategies such as \emph{Action Elimination}, which overcome this drawback by eliminating suboptimal arms based on observed feedback without requiring prior knowledge of $\Delta_i$.

\subsection{Action Elimination}
In this subsection, we analyse the Action Elimination (AE) algorithm, formally defined in Algorithm~\ref{Alg:act_elm}. The algorithm maintains two disjoint sets of arms: an \emph{active set} $\mathcal{A}_t$, containing arms that are not yet sufficiently explored, and an \emph{inactive set} $\mathcal{B}_t$, with $\mathcal{A}_t \cap \mathcal{B}_t = \varnothing$. Initially, $\mathcal{A}_t = [K]$ and $\mathcal{B}_t = \varnothing$. Let $\mathcal{L}_{\mathcal{A}_t}$ and $\mathcal{L}_{\mathcal{B}_t}$ denote the ordered lists of active and inactive arms, respectively. If an arm is removed from $\mathcal{A}_t$, the size of $\mathcal{L}_{\mathcal{A}_t}$ decreases but the relative ordering among the remaining active arms is preserved, while $\mathcal{L}_{\mathcal{B}_t}$ may be reordered when new arms are added. The overall ortdering $\mathcal{L}_t$ is formed by concatenating $\mathcal{L}_{\mathcal{A}_t}$ and $\mathcal{L}_{\mathcal{B}_t}$. To ensure that every active arm continues to receive exploration opportunities, $\mathcal{L}_{\mathcal{A}_t}$ is rotated in a round-robin manner across slots, whereas $\mathcal{L}_{\mathcal{B}_t}$ is maintained in ascending order of LCB values.  

While $|\mathcal{A}_t| > 1$ (active phase), the algorithm updates empirical error probabilities and their confidence intervals (LCB and UCB). Arms are eliminated from $\mathcal{A}_t$ and permanently added to $\mathcal{B}_t$ whenever their confidence intervals no longer overlap with others. Once only one arm remains active (commit phase), all arms are ordered by LCB values, and this final order is fixed for the rest of the horizon. Thus, AE progressively eliminates suboptimal arms while refining the cascade ordering.

Unlike the Explore and Commit (EC) algorithm, which explores uniformly for a fixed budget before committing, AE adaptively eliminates inferior arms as soon as enough evidence is gathered. This reduces unnecessary exploration and can yield tighter regret guarantees.


\begin{algorithm}	
	\SetAlgoLined
	\KwOut{Ordered list $\mathcal{L}_t$}
	Initialize $\mathcal{A}_1=\{1,2,\cdots,K\}$, $\mathcal{B}_1=\{\}$ $\mathcal{L}_1=$ random order of arms, $t=1$, $\mathcal{L}_{\mathcal{A}_t}=\mathcal{L}_1$, $\mathcal{L}_{\mathcal{B}_t}=()$, $S_i(0)=0$, $\hat{p}_i(0)=0$\\
	\For{$t\le T$}{
		Order arms as in list $\mathcal{L}_t$\\
		The cascade model shows the result of arm $I_t$ to the user \\
		Feedback of result $Y_{I_t}(t)$ is observed\\
		\For{$i\in[K]$}{
			$S_{i}(t)=S_{i}(t-1)+\mathbbm{1}_{\{i=I_t\}}$\\
            $\hat{p}_{i}(t)=\left\{\sum_{n=1}^t \mathbbm{1}_{\{i=I_n\}}(1-Y_{i}(t))\right\}/S_i(t)$\\ 
			$L_i(t)=\hat{p}_i(t)-\sqrt{\frac{2 \log T}{S_i(t)}}$\\
			$U_i(t)=\hat{p}_i(t)+\sqrt{\frac{2 \log T}{S_i(t)}}$
		}	
		\If{ $[L_i(t),U_i(t)]\cap [L_j(t),U_j(t)]=\phi, \forall i\ne j$}{
			$\mathcal{A}_{t+1}=\mathcal{A}_{t}/\{j\}$\\
			$\tilde{\mathcal{L}}_{\mathcal{A}_{t+1}}=\mathcal{L}_{\mathcal{A}_{t}}/\{j\}$\\
			$\mathcal{B}_{t+1}=\mathcal{B}_{t}\cup\{j\}$\\
		}
		\If{$|\mathcal{A}_{t+1}|>1$}{ 			
				$\mathcal{L}_{A_{t+1}}=(\tilde{l}_2^{A_t},\tilde{l}_3^{A_t},\cdots,\tilde{l}^{A_t}_{|\mathcal{A}_t|},\tilde{l}_1^{A_t})$
			
			$\mathcal{L}_{\mathcal{B}_{t+1}}=$ ascending order over $L_i(t)$, $i\in \mathcal{B}_{t+1}$\\		$\mathcal{L}_{t+1}=(\mathcal{L}_{\mathcal{A}_{t+1}},\mathcal{L}_{\mathcal{B}_{t+1}})$
		}
		\Else{$\mathcal{L}_{t+1}=$ ascending order over $L_i(t)$}	
		$t++$
	}	
	\caption{Action elimination}
	\label{Alg:act_elm}
\end{algorithm}



For the analysis, we introduce the following notations. Let $\Delta_i'=\min\{p_{i}-p_{i-1},\,p_{i+1}-p_{i}\}$ denote the minimum gap in error probability that distinguishes arm $i$ from its immediate neighbors. Define $N_i$ as the number of pulls required for arm $i$ to collect $\frac{16\log T}{\Delta_i'^2}$ feedback samples when placed at the head of the cascade, and let $N_i(t)$ denote the number of times arm $i$ has occupied the first position up to time $t$. Since Algorithm~\ref{Alg:act_elm} rotates arms in a round-robin fashion, each active arm is guaranteed opportunities to accumulate these samples. Let $T_s^{AE}$ represent the time at which the active phase ends. Finally, define the event $\mathcal{E}_t$ as 
\[
\mathcal{E}_t = \Big\{ |\hat{p}_i(t)-p_i| < \epsilon_i(t), \quad \forall i \Big\},
\]
where $\epsilon_{i}(t)=\sqrt{\tfrac{2\log T}{S_i(t)}}$ and $S_i(t)$ is the number of times arm $i$ has been observed up to time $t$.

\begin{theorem}
	The regret obtained by Algorithm \ref{Alg:act_elm} is 
	\begin{align*}
		\mathcal{R}^{AE}(T)= \OO(\log T).
	\end{align*}
\end{theorem}
\begin{proof}
	\begin{align*}
		\mathcal{R}^{AE}(T)\le & \sum_{t=1}^{T_s^{AE}} \E[\mathcal{R}_t]+\sum_{t=T_s^{AE}+1}^T \Prob(\mathcal{L}_t\ne \mathcal{L}^*)\tilde{\Delta}_{max}\\
		\overset{(a)}{\le} & \sum_{i=1}^K  \E[N_i]\tilde{\Delta}_i+ \frac{K\tilde{\Delta}_{max}}{T}+\frac{K\tilde{\Delta}_{max}}{T} \\
		\le & \sum_{i=1}^K \frac{16\log T}{\Delta_i'^2 \mu_i} \tilde{\Delta}_i +\frac{2K\tilde{\Delta}_{max}}{T}\\
		=& \OO(\log T),
	\end{align*}
	where $(a)$ is obtained by using Lemma \ref{lem:AE_confd} and Lemma \ref{lem:AE_act_bound}.
\end{proof}

\begin{lemma} \label{lem:AE_confd}
	The probability of Algorithm \ref{Alg:act_elm} choosing a sub-optimal ordering in the commit phase is bounded as follows
	\begin{align*}
		\Prob(\mathcal{L}_t\ne \mathcal{L}^*) \le \frac{K}{T^2}.
	\end{align*}
\end{lemma}

\begin{lemma}\label{lem:AE_act_bound}
	For Algorithm \ref{Alg:act_elm}, regret in active phase is bounded as follows
	\begin{align*}
		\sum_{t=1}^{T_s^{AE}} \E[\mathcal{R}_t]\le \sum_{i=1}^K  \E[N_i]\tilde{\Delta}_i+ \frac{K\tilde{\Delta}_{max}}{T}.
	\end{align*}
\end{lemma}
\begin{lemma} \label{lem:lb_eq_lim}
	Let $f(t)$ be a function of $t$ such that $0 \le f(t) \le t$. Then, for constants $\alpha_1, \alpha_2, \alpha_3 > 0$, we have  
	$$\alpha_1 f(t) + \alpha_2 e^{-\alpha_3 f(t)} \, (t-f(t)) = \Omega(\log t).$$
\end{lemma}
\noindent The proof of Lemma \ref{lem:AE_confd}, \ref{lem:AE_act_bound}, \ref{lem:lb_eq_lim} is provided in the Appendix \cite{techdoc}.
\begin{theorem}\label{thm:AE_lb}
	The regret of Algorithm \ref{Alg:act_elm} is lower bounded as
	\begin{align*}
		\mathcal{R}^{AE}(T) = \Omega(\log(T)).
	\end{align*}
\end{theorem}
\begin{proof}
	Let $\delta$ denote the probability that the algorithm commits to a sub-optimal ordering, and let $T_s^{AE}$ represent the number of rounds spent in the active phase before committing. 
	
	By results on best arm identification with fixed confidence in the full-information setting \cite{lattimore2020bandit}, the expected length of the active phase must satisfy
	\begin{align*}
		\E[T_s^{AE}] \;\ge\; \beta' \log\!\left(\tfrac{1}{\delta}\right),
	\end{align*}
	for some constant $\beta'>0$. 
	    
	An important property of Algorithm~\ref{Alg:act_elm} is that all arms in the active set $\mathcal{A}_t$ are explored in a round-robin manner through the orderings $\mathcal{L}_{\mathcal{A}_t}$. The algorithm remains in the active phase as long as at least two arms are active. Consequently, the maximum number of times any single ordering can be selected is $T_s^{AE}/2$, which implies that a sub-optimal ordering is chosen in at least half of the slots. This structural property of the algorithm directly contributes to the regret incurred during the active phase. Therefore,
	\begin{align*}
		\mathcal{R}(T_s^{AE}) 
		\;\ge\; \frac{\E[T_s^{AE}]}{2}\,\tilde{\Delta}_{\min}
		\;\ge\; \frac{1}{2}\beta'\log\!\left(\tfrac{1}{\delta}\right)\tilde{\Delta}_{\min}.
	\end{align*}
	
	After the active phase, if the commit phase chooses a sub-optimal ordering (which happens with probability at least $\delta$), regret continues to accumulate linearly with rate $\tilde{\Delta}_{\min}$. Thus, the total regret satisfies
	\begin{align}
		\mathcal{R}^{AE}(T)
		&\;\ge\; \mathcal{R}(T_s^{AE}) + \sum_{t=T_s^{AE}+1}^T \delta \,\tilde{\Delta}_{\min} \nonumber \\ 
		&\;\ge\; \frac{1}{2}\beta' \log\!\left(\tfrac{1}{\delta}\right)\tilde{\Delta}_{\min} 
		+ (T-T_s^{AE}-1)\delta\tilde{\Delta}_{\min}. \label{eq:lb_AE}
	\end{align}	
	Finally, applying Lemma \ref{lem:lb_eq_lim} to \eqref{eq:lb_AE} yields the claimed lower bound,
	$$\mathcal{R}^{AE}(T) = \Omega(\log(T)).$$
\end{proof}

	
	

\begin{remark}
    From the lower bound analysis, we can conclude that any policy that commits after a certain exploration period will necessarily incur a regret of order $\Omega(\log T)$.
\end{remark}

\subsection{LCB}
 In this subsection, we analysed the Lower Confidence Bound (LCB) algorithm, formally described in Algorithm~\ref{Alg:LCB}. The key idea of LCB is to order the arms at each round according to their estimated error probabilities, adjusted by a confidence term that encourages exploration. Initially, each arm is pulled until at least one user feedback is obtained to ensure a valid estimate. At every round $t$, the arms are ranked in ascending order of their lower confidence bounds $L_i(t)$, defined as the empirical error estimate $\hat{p}_i(t)$ reduced by a confidence margin $\sqrt{\tfrac{2\log t}{S_i(t)}}$, where $S_i(t)$ is the number of times feedback for arm $i$ has been observed. This construction balances exploration and exploitation by prioritising arms that either appear to have lower error probability or are still under-explored. Over time, the ordering of arms converges toward the optimal cascade order as the confidence intervals shrink with more observations.

\begin{algorithm}	
	\SetAlgoLined
	\KwOut{Ordered list $\mathcal{L}_t$}
	Pull each arm till at least one user feedback is obtained\\
	\While{$t\le T$}{
		$\mathcal{L}_t$= List arms in ascending order of $L_{i}(t)$\\	
		Observe user feedback: $I_t$\\
		$S_{I_t}(t)=S_{I_t}(t-1)+1$\\
		$\hat{p}_{I_t}(t)=\hat{p}_{I_t}(t-1)+\frac{1}{S_{I_t}(t)}(1-Y_{I_t}(t)-\hat{p}_{I_t}(t-1))$\\
		$L_i(t)=\hat{p}_i(t)-\sqrt{\frac{2\log t}{S_i(t)}}$, for all $i\in [K]$
	}
	
	\caption{LCB}
	\label{Alg:LCB}
\end{algorithm}

\begin{theorem}\label{thm:LCB}
	The regret obtained by Algorithm \ref{Alg:LCB} is 
	\begin{align*}
		\mathcal{R}^{LCB}(T)=\OO(1).
	\end{align*}
\end{theorem}
The proof of Theorem~\ref{thm:LCB} is provided in the Appendix \cite{techdoc}.

Theorem~\ref{thm:LCB} shows that the regret of the LCB algorithm is $\OO(1)$, 
which is significantly stronger than the $\Omega(\log T)$ lower bounds established 
for EC (Theorem~\ref{thm:EC_lb}) and AE (Theorem~\ref{thm:AE_lb}). It is worth noting that, in the standard stochastic bandit setting, both Explore-then-Commit and Action Elimination are known to achieve order-optimal regret guarantees. However, this is no longer the case in our problem, where the commitment inherent in these algorithms leads to $\Omega(\log T)$ regret. In contrast, the LCB algorithm avoids committing to a single arm, instead adapting continuously through confidence-bound updates, which allows it to achieve significantly lower regret in our setting.



\subsection{Thompson Sampling}
In this subsection, we analysed the Thompson Sampling (TS) algorithm, formally defined in Algorithm~\ref{Alg:TS}. TS is a Bayesian approach that maintains a posterior distribution over the error probability of each arm, modelled using Beta priors. At each round $t$, a sample $\theta_i(t)$ is drawn from the Beta distribution $\text{Beta}(\alpha_i(t),\beta_i(t))$ for each arm $i\in[K]$. The arms are then ordered in ascending order of these sampled values, thereby balancing exploration and exploitation through randomisation. After observing the user feedback $Y_{I_t}(t)$ for the triggered arm $I_t$, the corresponding posterior parameters are updated: $\alpha_{I_t}$ is incremented when the feedback indicates success, and $\beta_{I_t}$ is incremented otherwise. This sampling-based update mechanism ensures that arms with higher uncertainty are explored more frequently, while arms with consistently low error probabilities are more likely to appear earlier in the cascade, leading to convergence toward the optimal ordering.

\begin{algorithm}	
	\SetAlgoLined
	\KwOut{Ordered list $\mathcal{L}_t$}
	Initialize: $\alpha_i=1$, $\beta_i=1$ for all $i \in [K]$\\
	\While{$t\le T$}{
		Generate Thompson sample $\theta_i(t)\sim \text{Beta}(\alpha_i(t),\beta_i(t))$ for all $i\in [K]$\\
		$\mathcal{L}_t$= List arms in ascending order of $\theta_{i}(t)$\\	
		The cascade model shows the result of arm $I_t$ to the user \\
		Feedback of result $Y_{I_t}(t)$ is observed\\
		$\alpha_{I_t}(t+1)=\alpha_{I_t}(t)+1-Y_{I_t}(t)$, 	$\beta_{I_t}(t+1)=\beta_{I_t}(t)+Y_{I_t}(t)$.
	}	
	\caption{Thompson sampling}
	\label{Alg:TS}
\end{algorithm}

\begin{theorem}\label{thm:TS}
	The regret obtained by Algorithm~\ref{Alg:TS} satisfies
	\begin{align*}
		\mathcal{R}^{TS}(T)= \OO(1).
	\end{align*}
\end{theorem}
\begin{proof} 
	Regret is incurred at time slot $t$ if $I_t=i$ and there exists $k$ such that $p_i>p_k$ while arm $i$ appears ahead of arm $k$ in the cascade, i.e., $l^{-1}_t(i)<l^{-1}_t (k)$. Hence, we can upper bound the cumulative regret as
    \begin{align*}
		\mathcal{R}&(T)\\
		\le& \tilde{\Delta}_{max}.\\
		&\E\left[\sum_{t=1}^{T} \sum_{i=2}^K \Indc\{I_t=i, \exists k \text{ s.t } p_i>p_k, l^{-1}_t(i)<l^{-1}_t (k) \}\right].
	\end{align*}
	Define the event $A_{i,k}(t)$ as $\{I_t=i,\, l^{-1}_t(i)<l^{-1}_t (k),\, p_i>p_k\}$.  
	Let $E^{p}_{i,k}(t)$ denote the event $\{\hat{p}_i(t)<p_i-\Delta_{i,k}/4\}$ and $E^{\theta}_{i,k}(t)$ denote $\{\theta_{i}(t)<p_{k}+\Delta_{i,k}/4\}$, where $\Delta_{i,k}=p_i-p_k$.
	
	Therefore, the regret decomposition becomes
	\begin{align*}
		\mathcal{R}(T)
		&\le \tilde{\Delta}_{\max} \sum_{i=2}^{K}\sum_{k=1}^{K-1} 
		\E \Bigg[\sum_{t=1}^{T}  
		\Indc\{A_{i,k}(t),E^{p}_{i,k}(t)\} \\
		&\quad+ \Indc\{A_{i,k}(t),\bar{E}^{p}_{i,k}(t),E^{\theta}_{i,k}(t)\}\\
		&\quad + \Indc\{A_{i,k}(t),\bar{E}^{p}_{i,k}(t),\bar{E}^{\theta}_{i,k}(t)\}\Bigg]. \numberthis \label{eq:TS_reg_decomp}
	\end{align*}
	Finally, by applying Lemmas~\ref{lem:TS_term1}, \ref{lem:TS_term2}, and \ref{lem:TS_term3}, which respectively control the contributions of each of the three terms in \eqref{eq:TS_reg_decomp}, we conclude that the cumulative regret is bounded by a constant, i.e., $\mathcal{R}^{TS}(T)=\OO(1)$.
\end{proof}


\begin{lemma} \label{lem:TS_term1}
	The regret of the first term in \eqref{eq:TS_reg_decomp} is bounded as follows
	\begin{align*}
		\E\left[\sum_{t=1}^{T}\Indc\{A_{i,k}(t),E^{p}_{i,k}(t)\}\right]\le 1+\frac{16}{\Delta_{i,k}^2}.
	\end{align*}
\end{lemma}

\begin{lemma} \label{lem:TS_term2}
	The regret of the second term in \eqref{eq:TS_reg_decomp} is bounded as follows
	\begin{align*}
		\E\left[\sum_{t=1}^T\Indc\{A_{i,k}(t),\bar{E}^{p}_{i,k}(t),E^{\theta}_{i,k}(t)\}\right]\le \frac{32 }{\Delta_{i,k}^2\bar{\mu}_i} +\frac{2}{\bar{\mu}_i^2}+ \frac{\pi^2}{6}.
	\end{align*}
\end{lemma}

\begin{lemma} \label{lem:TS_term3}
	The regret of the third term in \eqref{eq:TS_reg_decomp} is bounded as follows
	\begin{align*}
		\E\Bigg[\sum_{t=1}^T&\Indc\{A_{i,k}(t),\bar{E}^{p}_{i,k}(t),\bar{E}^{\theta}_{i,k}(t)\}\Bigg]\\
		\le & \frac{24}{\epsilon^2}+ \frac{2c_1}{\epsilon^2} +\frac{c_1e^{\epsilon^2/2}}{\epsilon^2}\left( \frac{1}{16\epsilon}\Indc\{16\epsilon<1\}+\frac{1}{e}\right)\\ &+\frac{8c_1}{\epsilon}-\frac{4c_1\log (e^{2\epsilon}-1)}{\epsilon^2},
	\end{align*}
	where $c_1$ is a constant.
\end{lemma}
The proof of Lemma \ref{lem:TS_term1}, \ref{lem:TS_term2}, \ref{lem:TS_term3} are provided in the Appendix \cite{techdoc}.

From Theorem~\ref{thm:TS} we observe that Thompson Sampling also outperforms EC and AE in terms of regret minimization. Similar to LCB, Thompson Sampling avoids committing to a fixed arm after an exploration phase. Instead, it maintains a posterior distribution over arm parameters and samples from it at each round, thereby naturally balancing exploration and exploitation throughout the horizon. This probabilistic updating enables Thompson Sampling to achieve constant regret, whereas EC and AE incur $\Omega(\log T)$ regret due to their one-time commitment strategy.

\section{Simulations}
\begin{figure}[hbt]
\centering
\includegraphics[width=0.9\linewidth]{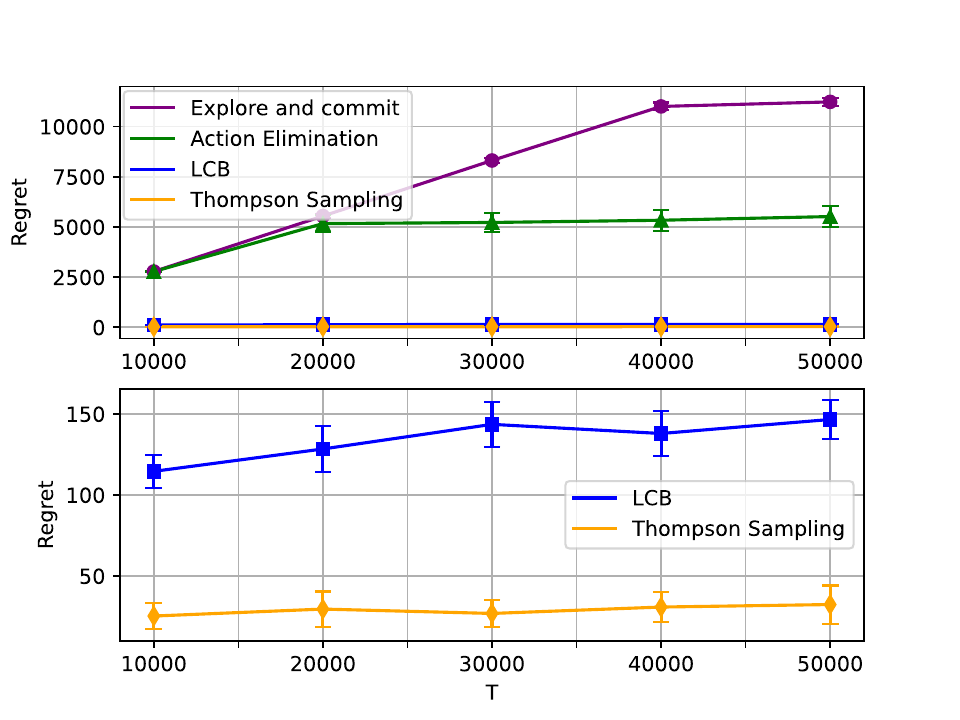}
\caption{Comparison of regret for different policies}
\label{fig:stch_reg}
\end{figure}

In this section, we validate our theoretical findings through simulations. We consider the case of $K=5$ arms with parameters $\mu = [0.85,0.9,0.95,0.92,0.87]$ and $p = [0.1,0.25,0.4,0.55,0.7]$. The results are averaged over 20 independent experiments. As shown in \figurename~\ref{fig:stch_reg}, we compare the regret of the policies from Section~\ref{sec:results} for varying horizon $T$. The results indicate that the Thompson Sampling and LCB algorithms achieve constant regret, whereas the Action Elimination and Explore-and-Commit algorithms exhibit logarithmic regret growth. These observations are consistent with, and hence validate, the theoretical guarantees established in Section~\ref{sec:results}. The superior performance of Thompson Sampling and LCB arises from their ability to continuously balance exploration and exploitation, thereby adapting to uncertainty throughout the horizon, while Action Elimination and Explore-and-Commit commit after the exploration period, which inherently leads to $\Omega(\log T)$ regret. To ensure fair initialisation, we assign a large value to the LCB at the beginning and break ties randomly, thereby avoiding additional delays for initial sampling.

\begin{figure}[hbt]
\centering
\includegraphics[width=0.9\linewidth]{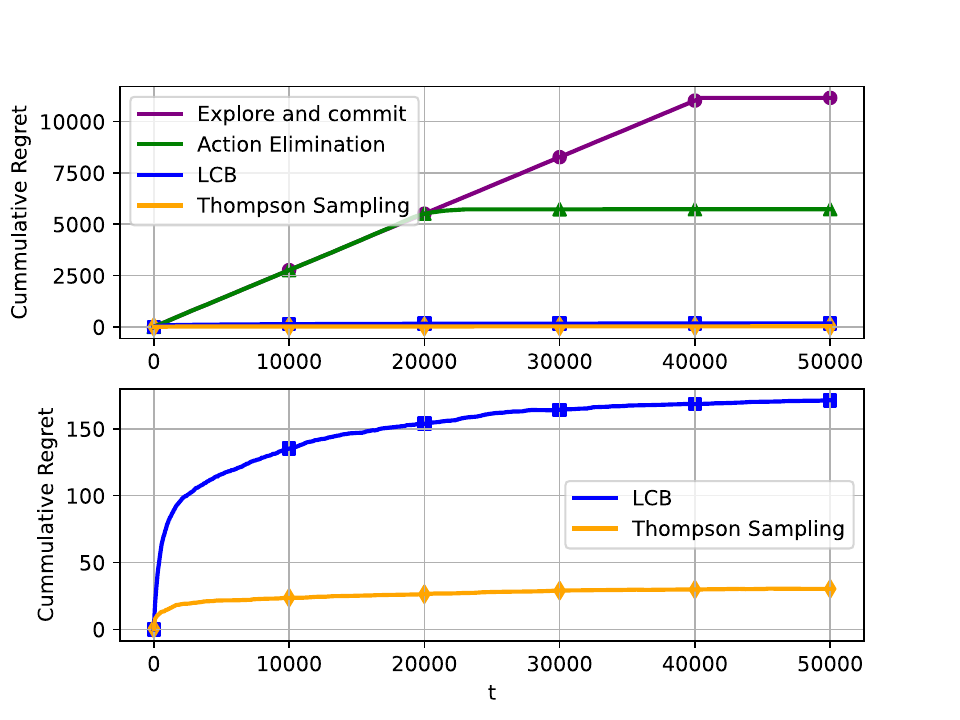}
\caption{Comparison of cumulative regret of different polices}
\label{fig:stch_reg2}
\end{figure}

To analyze how regret evolves over time and how frequently each policy selects a suboptimal ordering, we plot the cumulative regret up to $T=5\times 10^4$ rounds for the case of $K=5$ arms with parameters $\mu = [0.85, 0.9, 0.95, 0.92, 0.87]$ and $p = [0.1, 0.25, 0.4, 0.55, 0.7]$. The results are presented in \figurename~\ref{fig:stch_reg2}. We observe that the Explore-and-Commit and Action Elimination algorithms incur constant regret once they enter the commit phase. In contrast, the LCB and Thompson Sampling algorithms continue to accumulate regret, but at a much slower rate. While they initially incur small regret due to exploration, their regret growth eventually saturates and becomes sublinear, highlighting their superior ability to balance exploration and exploitation over time.

\newpage

\section*{References}
\bibliography{ref_PC}
\bibliographystyle{IEEEtran}

\footnote{AI tools are used throughout the paper for grammar and editing.}
 \section{Appendix}
\begin{proof}[Proof of Theorem \ref{thm:opt_policy}]
We prove this result by contradiction. Assume that the optimal arm ordering, denoted by $\mathcal{L}^*$, is not sorted by increasing $p_i$ values. This implies there exists an adjacent pair of arms, $l_i^*$ and $l_{i+1}^*$, such that $p_{l_i^*} > p_{l_{i+1}^*}$.

The expected reward of the assumed optimal ordering $\mathcal{L}^*$ is given by
	\begin{align*}
		r^*=\sum_{i=1}^{K}(1-p_{l_i^*})\mu_{l_i^*}\prod_{j=1}^{i-1}(1-\mu_{l_j^*}).
	\end{align*}
Now, let's consider a new ordering, $\tilde{\mathcal{L}}$, obtained by swapping arms $l_i^*$ and $l_{i+1}^*$. The difference in expected reward between the two orderings is
\begin{align*}
r^* - \tilde{r} &= \left( \prod_{j=1}^{i-1} (1-\mu_{l_j^*}) \right) \left[ (1-p_{l_i^*})\mu_{l_i^*} + (1-p_{l_{i+1}^*})\mu_{l_{i+1}^*}(1-\mu_{l_i^*}) \right. \\
&\quad \left. - \left( (1-p_{l_{i+1}^*})\mu_{l_{i+1}^*} + (1-p_{l_i^*})\mu_{l_i^*}(1-\mu_{l_{i+1}^*}) \right) \right] \\
&= \left( \prod_{j=1}^{i-1} (1-\mu_{l_j^*}) \right) \left[ (1-p_{l_i^*})\mu_{l_i^*}\left(1-\left(1-\mu_{l_{i+1}^*}\right) \right) \right. \\
&\quad \left. -  (1-p_{l_{i+1}^*})\mu_{l_{i+1}^*}\left(1-\left(1-\mu_{l_i^*}\right) \right) \right] \\
&= \left( \prod_{j=1}^{i-1} (1-\mu_{l_j^*}) \mu_{l_i^*}\mu_{l_{i+1}^*} \right) \left[ (1-p_{l_i^*}) - (1-p_{l_{i+1}^*}) \right] \\
&= \left( \prod_{j=1}^{i-1} (1-\mu_{l_j^*}) \mu_{l_i^*}\mu_{l_{i+1}^*} \right) (p_{l_{i+1}^*} - p_{l_i^*}).
\end{align*}
Since probabilities are non-negative, the term $\left( \prod_{j=1}^{i-1} (1-\mu_{l_j^*}) \mu_{l_i^*}\mu_{l_{i+1}^*} \right)$ is non-negative. From our assumption, $p_{l_i^*} > p_{l_{i+1}^*}$, which implies $p_{l_{i+1}^*} - p_{l_i^*} < 0$. Therefore, $r^* - \tilde{r} < 0$, which means $\tilde{r} > r^*$. Thus $\tilde{r} > r^*$, contradicting the optimality of $\mathcal{L}^*$. Therefore no such index $i$ can exist, which proves the claimed ordering $p_{l_1^*}<p_{l_2^*}<\cdots <p_{l_K^*}$.

\end{proof}

\begin{proof}[Proof of Lemma \ref{lem:lb_eq_lim}]
We analyse the asymptotic behaviour of 
\[
\frac{\alpha_1 f(t)+\alpha_2 e^{-\alpha_3 f(t)} (t-f(t))}{\log(t)}
\]
under different growth rates of $f(t)$.

\noindent\textbf{Case 1:} If $\liminf_{t\to\infty} \tfrac{f(t)}{\log(t)} = \infty$, then
\begin{align*}
    \liminf_{t\to\infty} \frac{\alpha_1 f(t)+\alpha_2 e^{-\alpha_3 f(t)} (t-f(t))}{\log(t)}
\;&\ge\; \liminf_{t\to\infty} \frac{\alpha_1 f(t)}{\log(t)}\\
&=\infty.
\end{align*}
\noindent\textbf{Case 2:} If $\liminf_{t\to\infty} \tfrac{f(t)}{\log(t)}=\ell<\infty$, then
\begin{align*}
&\liminf_{t\to\infty} \frac{\alpha_1 f(t)+\alpha_2 e^{-\alpha_3 f(t)} (t-f(t))}{\log(t)} \\
&= \alpha_1 \ell + \alpha_2 \liminf_{t\to\infty} 
t^{-\alpha_3 f(t)/\log(t)} \Big(\tfrac{t}{\log(t)}-\ell\Big).
\end{align*}
Since $\liminf_{t\to\infty} \tfrac{f(t)}{\log(t)}=\ell$, we obtain
\begin{align*}
    \liminf_{t\to\infty}\frac{\alpha_1 f(t)+\alpha_2 e^{-\alpha_3 f(t)} (t-f(t))}{\log(t)}\\ =
\begin{cases}
\alpha_1 \ell, & \alpha_3\ell>1,\\[0.5ex]
\alpha_1 \ell+\alpha_2, & \alpha_3\ell=1,\\[0.5ex]
\infty, & \alpha_3\ell<1.
\end{cases}
\end{align*}

\noindent Combining both cases, the quantity is always bounded below by a positive constant multiple of $\log(t)$. Hence, 
\[
\alpha_1 f(t)+\alpha_2 e^{-\alpha_3 f(t)} (t-f(t)) = \Omega(\log(t)).
\]
\end{proof}

\color{black}

\begin{proof}[Proof of Theorem \ref{thm:EC_lb}]
	Let $N_i(t)$ denote the number of times arm $i$ is placed first in the cascade up to time $t$. 
	Let $\delta(T_s^{EC})$ represent the probability that Algorithm~\ref{Alg:Exp_cmmt} chooses a suboptimal ordering in the commit phase after the exploration horizon $T_s^{EC}$. 
	
	\textbf{Exploration phase:}  
	Since the algorithm rotates arms uniformly, each arm appears first approximately $T_s^{EC}/K$ times.  
	The regret incurred during this phase comes from pulling suboptimal arms in the first position, and can be written as
	\[
	\mathcal{R}^{EC}(T_s^{EC}) \;\ge\; \sum_{i=2}^{K} \E[N_i(T_s^{EC})] \tilde{\Delta}_{\min}
		= \frac{K-1}{K} T_s^{EC} \tilde{\Delta}_{\min}.
	\]

	\textbf{Commit phase:}  
	If a suboptimal ordering is chosen after exploration, then the regret in the commit phase is at least 
	\[
	\delta(T_s^{EC})(T - T_s^{EC} - 1)\tilde{\Delta}_{\min}.
	\]
     During exploration, the algorithm collects at most $T_s^{EC}$ effective samples of arms in the first position. Thus, the problem of finding the optimal cascade ordering contains, as a subproblem, best-arm identification with a fixed budget of $T_s^{EC}$ samples. Therefore, any lower bound on the error probability of fixed-budget best-arm identification directly applies to our setting. By the result of \cite{audibert2010best}, there exists a constant $\beta>0$ such that
	$$\delta(T_s^{EC}) \;\ge\; e^{-\beta T_s^{EC}}.$$
    This inequality means that the probability of choosing a suboptimal ordering in the commit phase cannot be made arbitrarily small. Even after exploring each arm for $T_s^{EC}$ rounds, there is still a nonzero chance that the algorithm misidentifies the best arm for the first position. Therefore, the total regret satisfies
	\begin{align*}
	\mathcal{R}^{EC}(T) 
	&\ge \mathcal{R}^{EC}(T_s^{EC}) + \delta(T_s^{EC})(T - T_s^{EC} - 1)\tilde{\Delta}_{\min} \\
	&\ge \frac{K-1}{K} T_s^{EC} \tilde{\Delta}_{\min} 
		+ e^{-\beta T_s^{EC}}(T - T_s^{EC} - 1)\tilde{\Delta}_{\min}.
	\end{align*}
	Finally, by Lemma~\ref{lem:lb_eq_lim}, this simplifies to
	\[
		\mathcal{R}^{EC}(T) = \Omega(\log T).
	\]
\end{proof}

\begin{proof}[Proof of Lemma \ref{lem:EC_reg}]
	The arms' ordering is not optimal if the LCBs are not ordered correctly. Which means $\exists i$ such that $L_{i}(t)>L_{i+1}(t)$, $t=T^{EC}_s$. 
	\begin{align*}
		\Prob(\mathcal{L}_t\ne \mathcal{L}^*) \le \sum_{i=1}^{K-1} \Prob(L_{i}(t)>L_{i+1}(t)).
	\end{align*}
	Let $\mathcal{E}_t$ be the event that $|\hat{p}_{i}(t)-p_i|<\sqrt{\frac{2\log T}{S_{i}(t)}}$ for all $i$.
	\begin{align*}
		\Prob(\mathcal{L}_t &\ne \mathcal{L}^*)\\
		\le& \sum_{i=1}^{K-1} \Prob(L_{i}(t)>L_{i+1}(t),\mathcal{E}_t)+ \Prob(\mathcal{E}_t^c)\\
		\overset{(a)}{\le}&\sum_{i=1}^{K-1}\Prob\left(\hat{p}_{i}(t)-\sqrt{\frac{2\log T}{S_{i}(t)}}> \hat{p}_{i+1}(t)-\sqrt{\frac{2\log T}{S_{i+1}(t)}}, \mathcal{E}_t\right)\\
		&+\frac{K}{T^4}\\
		\le&\sum_{i=1}^{K-1}\Prob\left(p_{i}> {p}_{i+1}-2\sqrt{\frac{2\log T}{S_{i+1}(t)}}\right)+\frac{K}{T^4}\\
		= & \sum_{i=1}^{K-1} \Prob\left(S_{i+1}(t)<\frac{8 \log T}{\Delta_{i+1}^2}\right)+\frac{K}{T^4}\\
		= & \sum_{i=2}^{K} \Prob\left(S_{i}(T_s^{EC})<\frac{8 \log T}{\Delta_{i}^2}\right)+\frac{K}{T^4}\\
		\le& \sum_{i=2}^K \Prob\left(\sum_{\tau=1}^{T^{EC}_s}X_{i}(\tau) \Indc\{l_1^{(\tau)}=i\} <\frac{8 \log T}{\Delta_i^2}\right)+\frac{K}{T^4}\\
		\le&\sum_{i=2}^K \Prob\left(\sum_{\tau=1}^{T^{EC}_s}X_{i}(\tau) \Indc\{l_1^{(\tau)}=i\} <N \mu_i/2\right)+\frac{K}{T^4}\\
		\overset{(b)}{\le} & \sum_{i=2}^K e^{-N\mu_i/8}+ \frac{K}{T^4}\\
		\le & \sum_{i=2}^K e^{-n_i\mu_i/8}+ \frac{K}{T^4}\\
		\le &\sum_{i=2}^K\frac{1}{T^{2/\Delta^2_i}}+\frac{K}{T^4}\\
		\le &\frac{K}{T^{2}}+\frac{K^2}{T^4}.
	\end{align*}
	 Where $(a)$ is obtained using Hoeffding's inequality, $(b)$ is obtained by using the fact that if $X\sim Ber(n,p)$ and $\E[X]=\mu$, then for $0<\epsilon<1$, $\Prob(X<(1-\epsilon)\mu)\le e^{-\epsilon^2\mu/2}$. 
\end{proof}

\begin{lemma}\label{lem:AE_opt}
	If $\mathcal{E}_t$ holds then in commit phase for Algorithm \ref{Alg:act_elm}, 
	$\mathcal{L}_t=\mathcal{L}^*$.
\end{lemma}
\begin{proof}
	Let us assume $\mathcal{E}_t$ holds and $\mathcal{L}_t\ne\mathcal{L}^*$ that is $\exists i$ in the ordering $\mathcal{L}_t$ such that $p_{l_i^t}<p_{l_{i-1}^t}$. Algorithm \ref{Alg:act_elm} is in commit phase.
	\begin{align*}
		\implies \hat{p}_{l_i^t}(t)-\epsilon_{l_{i}^t}(t) >& \hat{p}_{l_{i-1}^t}(t)+\epsilon_{l_{i-1}^t}(t)\\
		\implies p_{l_i^t}>&p_{l_{i-1}^t},
	\end{align*}
	which is a contradiction.
\end{proof}

\begin{proof}[Proof of Lemma \ref{lem:AE_confd}] 
	\begin{align*}
		\Prob(\mathcal{L}_t\ne \mathcal{L}^*)=& \Prob(\mathcal{L}_t\ne \mathcal{L}^*,\mathcal{E}_t)+\Prob(\mathcal{L}_t\ne \mathcal{L}^*,\mathcal{E}_t^c)\\
		\overset{(a)}{=}&\Prob(\mathcal{L}_t\ne \mathcal{L}^*,\mathcal{E}_t^c)\\
		\le & \Prob(\mathcal{E}_t^c)\\
		\le& \frac{K}{T^2},
	\end{align*}	
	where $(a)$ is obtained by using Lemma \ref{lem:AE_opt}.
\end{proof}

\begin{lemma}\label{lem:act_phase}
	For Algorithm \ref{Alg:act_elm}, if $\mathcal{E}_t$ holds and $S_{i}(t) > \frac{16\log T}{\Delta_i'^2}$ for all $i$ where, $\Delta_i'=\min\{p_{i}-p_{i-1},p_{i+1}-p_{i}\}$, then algorithm is not in the active phase.
\end{lemma}
\begin{proof}
	Let us assume $\mathcal{E}_t$ holds and algorithm is in active phase then $\exists i,j, k$  such that 
	\begin{align*}
		\hat{p}_i(t)+\epsilon_i(t)&>\hat{p}_{j}(t)-\epsilon_{j}(t) \text{ or }\\ \hat{p}_i(t)-\epsilon_i(t)&<\hat{p}_{k}(t)+\epsilon_{k}(t)\\
		\implies {p}_i+2\epsilon_i(t)&>{p}_{j}-2\epsilon_{j}(t) \text{ or } {p}_i-2\epsilon_i(t)<{p}_{k}+2\epsilon_{k}(t)\\
		\implies {p}_i+2\epsilon_i(t) &\ge \frac{p_{j}+p_i}{2} \text{ or } \frac{p_{j}+p_i}{2}\ge {p}_{j}-2\epsilon_{j}(t) \text{ or }\\
		{p}_i-2\epsilon_i(t)&\le \frac{p_{i}+p_{k}}{2} \text{ or }\frac{p_{i}+p_{k}}{2}\le {p}_{k}+2\epsilon_{k}(t)\\
		\implies		2\epsilon_{i}(t)\ge &\frac{p_{j}-p_i}{2} \text{ or }	2\epsilon_{j}(t)\ge \frac{p_{j}-p_i}{2} \text{ or}\\
		2\epsilon_{i}(t)\ge& \frac{p_{i}-p_{k}}{2} \text{ or } 2\epsilon_{k}(t)\ge \frac{p_{i}-p_{k}}{2}\\		
		\implies S_{i}(t)\le& \frac{16\log T}{(p_{j}-p_{i})^2} \text{ or } S_{j}(t)\le \frac{16\log T}{(p_{j}-p_{i})^2} \text{ or }\\
		S_{i}(t)\le& \frac{16\log T}{(p_{i}-p_{k})^2} \text{ or } S_{k}(t)\le \frac{16\log T}{(p_{i}-p_{k})^2}. \numberthis \label{eq:act_arm}
	\end{align*}
	Therefore if $S_{i}(t) > \frac{16\log T}{\Delta_i'^2}$ for all $i$ where, $\Delta_i'=\min\{p_{i}-p_{i-1},p_{i+1}-p_{i}\}$, then $\nexists j,k$ such that $\hat{p}_i(t)+\epsilon_i(t)>\hat{p}_{j}(t)-\epsilon_{j}(t) \text{ or } \hat{p}_i(t)-\epsilon_i(t)<\hat{p}_{k}(t)+\epsilon_{k}(t)$ for all $i$. This means no active arms exist, and Algorithm \ref{Alg:act_elm} is in the commit phase.
\end{proof}

\begin{proof}[Proof of Lemma \ref{lem:AE_act_bound}]
	Let us consider an algorithm $\tilde{\mathcal{A}}$, where samples are updated only when it is the head of the cascade and everything is the same as Algorithm \ref{Alg:act_elm}. Let $T_s^{AE}$ and $\tilde{T_s^\mathcal{A}}$ be the time after which Algorithm \ref{Alg:act_elm} and $\tilde{\mathcal{A}}$ enter commit phase. Since samples are updated less often, $\tilde{A}$ takes more time to enter the commit phase; therefore, $T_s^{AE}<\tilde{T}_s^\mathcal{A}$.

	Note that both algorithms modify the ordering similarly (round robin). Therefore, at any given time, $N_i(t), t\le T_s^{AE}$ is the same for both algorithms. Since $N_i(t)$ is a monotone function $N_i(T_s^{AE})\le N_i(\tilde{T}_s^\mathcal{A})$.
	
	If $\mathcal{E}_t$ holds $\forall t$, then Lemma \ref{lem:act_phase} also holds and arm $i$ is removed from active set if $S_{i}(t) > \frac{16\log T}{\Delta_i'^2}$. Then $N_i$ represents the upper bound on the number of times arm $i$ is head of the cascade in active phase for $\tilde{\mathcal{A}}$. Thus $N_i(T_s^{AE})\le N_i(\tilde{T}_s^\mathcal{A}) \le N_i$ when $\mathcal{E}_t$ occurs $\forall t$. 
	\begin{align*}
		\sum_{t=1}^{T^{AE}_s}\E[R_t]=&\sum_{t=1}^{T^{AE}_s}\E[R_t|\mathcal{E}_t]\Prob(\mathcal{E}_t)+\E[R_t|\mathcal{E}^c_t]\Prob(\mathcal{E}^c_t)\\
		\le&\sum_{t=1}^{T^{AE}_s}\E[R_t|\mathcal{E}_t] +\tilde{\Delta}_{max} \Prob(\mathcal{E}^c_t)\\
		\le&\sum_{i=1}^K \E[N_i(T^{AE}_s)]\tilde{\Delta}_i+ \sum_{t=1}^{T^{AE}_s} \Prob(\mathcal{E}^c_t) \tilde{\Delta}_{max}\\
		\overset{(a)}{\le}&\sum_{i=1}^K \E[N_i]\tilde{\Delta}_i+\frac{K\tilde{\Delta}_{max}}{T},
	\end{align*}	
	where $(a)$ is obtained by using Lemma \ref{lem:AE_confd}.
\end{proof}

\begin{proof}[Proof of Theorem \ref{thm:LCB}]
	Let $\mathcal{E}_t$ be the event that $|\hat{p}_{i}(t)-p_i|<\sqrt{\frac{2\log t}{S_{i}(t)}}$ for all $i$. By Hoeffding's inequality  we have $\Prob(\mathcal{E}_t^c)\le \frac{K}{t^2}$. \color{black}
	Let $\mathcal{G}_t$ be the event that all arms are ordered correctly in time slot $t$. Thus $\mathcal{G}^c_t$ represents the event that there $\exists i\in [K]$ such that $L_{i}(t)>L_{i+1}(t)$. Therefore,
	\begin{align*}
		\Prob(\mathcal{G}_t^c,\mathcal{E}_t)\le&\sum_{i=1}^{K-1}\Prob(L_{i}(t)>L_{i+1}(t),\mathcal{E}_t)\\
		=&\sum_{i=1}^{K-1}\Prob(\hat{p}_i(t)-\epsilon_i(t)>\hat{p}_{i+1}(t)-\epsilon_{i+1}(t),\mathcal{E}_t)\\
		\le&\sum_{i=1}^{K-1}\Prob(p_i>{p}_{i+1}-2\epsilon_{i+1}(t))\\
		=&\sum_{i=1}^{K-1}\Prob\left(\epsilon_{i+1}(t)>\frac{p_{i+1}-p_{i}}{2}\right)\\
		=& \sum_{j=2}^{K}\Prob\left( S_{j,t} <\frac{8 \log t}{\Delta_j^2}\right).
	\end{align*}
	Let us define a new random variable 
	\begin{align*}
		Z_{i}(t)&=\begin{cases}
			1 & \text{if } X_{i}(t)=1 \text{ and } X_{j}(t)=0, \forall j\ne i, \\
			0 & \text{otherwise}.
		\end{cases}
	\end{align*}
	Note that arrivals are independent therefore $\{Z_{i}(t)\}_{t\ge1}$ are also independent across time. Note that $Z_{i}(t)\sim Ber(\mu_i\prod_{j\ne i}(1-\mu_j))$ and let us define $\bar{\mu}_i=\E[Z_{i}(t)]=\mu_i\prod_{j\ne i}(1-\mu_j)$. The number of user feedback samples is lower bounded as follows $ \sum_{n=1}^t Z_{i}(t)\le S_{i}(t)$. Therefore 
	\begin{align*}
		\Prob(\mathcal{G}^c_t,\mathcal{E}_t)\le \sum_{j=2}^{K}\Prob\left( Z_{i}(t) <\frac{8 \log t}{\Delta_j^2}\right).
	\end{align*}
	The regret is obtained only when the ordering is incorrect. Let $\E[{R}_t]$ be the regret incurred in time slot t.
	\begin{align*}
		\E[{R}_t]=&	\E[{R}_t|\mathcal{G}_t]\Prob(\mathcal{G}_t)+ \E[{R}_t|\mathcal{G}_t^c]\Prob(\mathcal{G}_t^c)\\
		=& \E[{R}_t|\mathcal{G}_t^c]\Prob(\mathcal{G}_t^c)\\
		= & \E[{R}_t|\mathcal{E}_t,\mathcal{G}^c_t]\Prob(\mathcal{G}^c_t,\mathcal{E}_t)+ \E[{R}_t|\mathcal{E}_t^c,\mathcal{G}^c_t]\Prob(\mathcal{G}^c_t,\mathcal{E}_t^c)\\
		\le& \tilde{\Delta}_{max} \sum_{j=2}^{K}\Prob\left( Z_{i}(t) <\frac{8 \log t}{\Delta_j^2}\right)+\tilde{\Delta}_{max}\Prob(\mathcal{E}^c_t).
	\end{align*}
	The overall regret is bounded as follows
	\begin{align*}
		\mathcal{R}^{LCB}&(T)\\
		&=\sum_{t=1}^T \E[{R}_t]\\
		&\le \sum_{t=1}^T \tilde{\Delta}_{max} \sum_{j=2}^{K}\Prob\left( Z_{i}(t) <\frac{8 \log t}{\Delta_j^2}\right)+\tilde{\Delta}_{max}\Prob(\mathcal{E}^c_t)\\
		&\le \tilde{\Delta}_{max}\sum_{j=2}^K\sum_{t=1}^T\Prob\left( Z_{i}(t) <\frac{8 \log t}{\Delta_j^2}\right)+\sum_{t=1}^T\tilde{\Delta}_{max}\frac{K}{t^2}.
	\end{align*}
	Note that for $t\ge T'$, where $T'=\frac{16}{\Delta_j^2\bar{\mu}_j}\left( \log \frac{16}{\Delta_j^2\bar{\mu}_j}\right)^2 $ we have $\frac{8\log t}{\Delta_j^2}< \frac{\bar{\mu}_j t}{2}$, therefore
	\begin{align*}
		\sum_{t=1}^T\Prob\left( Z_{i}(t) \le\frac{8 \log t}{\Delta_j^2}\right) \le& T'+\sum_{t=T'}^T \Prob\left( Z_{i}(t) <\frac{t\bar{\mu}_{j}}{2}\right)\\
		\le & T'+\sum_{t=T'}^T e^{-t\bar{\mu}_j^2/2}\\
		\le & T'+\frac{2}{\bar{\mu}_j^2}.
	\end{align*}
	Now, we bound the regret as follows
	\begin{align*}
		\mathcal{R}^{LCB}(T) \le& \tilde{\Delta}_{max}\sum_{j=2}^K\left( T'+\frac{2}{\bar{\mu}_j^2}\right)+\tilde{\Delta}_{max} \frac{K\pi^2}{6} \\
		=&\OO(1).
	\end{align*}
\end{proof}

\begin{proof}[Proof of Lemma \ref{lem:TS_term1}]
	Let $\tau_0=0$ and $\tau_1, \tau_2,\cdots $ be the time slots in which  sample for $p_i$ is obtained i.e. $I_t=i$. 
	\begin{align*}
		\E\left[\sum_{t=1}^{T}\Indc\{A_{i,k}(t),E^{p}_{i,k}(t)\}\right]\le& \E\left[\sum_{t=1}^T\Indc\{E^{p}_{i,k}(t), I_t=i\}\right]\\
		\le & \E\left[\sum_{k=0}^{T}\Indc\{E^{p}_{i,k}(\tau_k)\}\right]\\
		\overset{(a)}{\le}& 1+\sum_{k=1}^{T}e^{-2k\epsilon^2}\\
		\le & 1+\frac{1}{2\epsilon^2}\\
		=& 1+\frac{8}{\Delta_{i,k}^2},
	\end{align*}
	where $(a)$ is obtained by using Hoeffdings' inequality.
\end{proof}

\begin{lemma} \label{lem:lb_samples}
	If $S_i(t)$ represents the number of samples observed by arm $i$ till time $t$ then we have,
	\begin{align*}
		\sum_{t=1}^T \Prob \left( S_i(t)\le \frac{16\log t}{\Delta^2_{i,k}} \right) \le \frac{32 }{\Delta_{i,k}^2\bar{\mu}_i}+\frac{2}{\bar{\mu}_i^2}.
	\end{align*}
\end{lemma}
\begin{proof}
	Let us define a new random variable 
	\begin{align*}
		Z_{i}(t)=\begin{cases}
			1 & \mbox{if } X_{i}(t)=1 \mbox{ and } X_{j}(t)=0, \forall j\ne i, \\
			0 & \mbox{otherwise}.\\
		\end{cases}
	\end{align*}
	Note that arrivals are independent therefore $\{Z_{i}(t)\}_{t\ge1}$ are also independent across time. Note that $Z_{i}(t)\sim Ber(\mu_i\prod_{j\ne i}(1-\mu_j))$ and let us define $\bar{\mu}_i=\E[Z_{i}(t)]=\mu_i\prod_{j\ne i}(1-\mu_j)$. The number of user feedback samples is lower bounded as follows $ \sum_{n=1}^t Z_{i}(n)\le S_{i}(t)$. Therefore,
	
	\begin{align*}
		\sum_{t=1}^T\Prob\left(S_i(t)\le\frac{16\log t}{\Delta_{i,k}^2}\right) \le  
		\sum_{t=1}^T\Prob\left(\sum_{n=1}^t Z_{i}(n)\le \frac{16\log t}{\Delta_{i,k}^2} \right)
	\end{align*}
	Note that for $t\ge T'$, where $T'=\frac{32}{\Delta_{i,k}^2\bar{\mu}_i}$ we have $\frac{16\log t}{\Delta_{i,k}^2}< \frac{\bar{\mu}_j t}{2}$, therefore
	\begin{align*}
		\sum_{t=1}^T\Prob\left( Z_{i}(t) \le\frac{16 \log t}{\Delta_{i,k}^2}\right) \le& \frac{32 }{\Delta_{i,k}^2\bar{\mu}_i}+\sum_{t=T'}^T \Prob\left( Z_{i}(t) <\frac{t\bar{\mu}_{i}}{2}\right)\\
		\le & \frac{32 }{\Delta_{i,k}^2\bar{\mu}_i}+\sum_{t=T'}^T e^{-t\bar{\mu}_i^2/2}\\
		\le & \frac{32 }{\Delta_{i,k}^2\bar{\mu}_i}+\frac{2}{\bar{\mu}_i^2}.
	\end{align*}
\end{proof}

\begin{proof}[Proof of Lemma \ref{lem:TS_term2}]
Let $L_i(t)=\frac{16\log t}{\Delta_{i,k}^2}$ then,
	\begin{align*}
		\sum_{t=1}^T&\Indc\{A_{i,k}(t),\bar{E}^{p}_{i,k}(t),E^{\theta}_{i,k}(t)\}\\
		=& \sum_{t=1}^T\Indc\{A_{i,k}(t),\bar{E}^{p}_{i,k}(t),E^{\theta}_{i,k}(t), S_i(t)\le L_i(t)\}\\
		&+\sum_{t=1}^T\Indc\{A_{i,k}(t),\bar{E}^{p}_{i,k}(t),E^{\theta}_{i,k}(t), S_i(t)>L_i(t)\}.		
	\end{align*}
	Consider,
	\begin{align*}
		\E&\left[\sum_{t=1}^T\Indc\{A_{i,k}(t),\bar{E}^{p}_{i,k}(t),E^{\theta}_{i,k}(t), S_i(t)\le L_i(t)\}\right]\\
		\le &\sum_{t=1}^T\Prob(S_i(t)\le L_i(t))\\
		\overset{(b)}{\le}& \frac{32 }{\Delta_{i,k}^2\bar{\mu}_i} + \frac{2}{\bar{\mu}_i^2}, \numberthis \label{eq:thmp_term2_1}
	\end{align*}
	where $(b)$ is obtained from Lemma \ref{lem:lb_samples}.
	Now consider,
	\begin{align*}
		\E &\left[\sum_{t=1}^T\Indc\{A_{i,k}(t),\bar{E}^{p}_{i,k}(t),E^{\theta}_{i,k}(t), S_i(t)> L_i(t)\}\right]\\
		\le& \sum_{t=1}^T\E\left[\Indc\{\hat{p}_i(t)\ge p_i-\epsilon,\theta_{i}(t)<p_{k}+\epsilon, S_i(t)> L_i(t)\}\right]\\
		\le& \sum_{t=1}^T\E\left[\Indc\left(\theta_{i}(t)\le\hat{p}_i(t)-\Delta_{i,k}+2\epsilon, S_i(t)> \frac{16\log t}{\Delta_{i,k}^2}\right) \right]\\
		= & \sum_{t=1}^T\E\left[\Indc\left(\theta_{i}(t)\le\hat{p}_i(t)-\frac{\Delta_{i,k}}{2}, S_i(t)> \frac{16\log t}{\Delta_{i,k}^2}\right) \right]\\
		\le & \sum_{t=1}^T\Prob\left(\theta_{i}(t)\le\hat{p}_i(t)-\sqrt{ \frac{4\log t}{S_i(t)}}\right)\\
		\overset{(c)}{\le}& \sum_{t=1}^T\frac{1}{t^2}, \numberthis \label{eq:thmp_term2_2}
	\end{align*}
	where $(c)$ is obtained by using Lemma 4 of \cite{wang18a}. By using \eqref{eq:thmp_term2_1} and \eqref{eq:thmp_term2_2}, we get the result stated.
\end{proof}

\begin{proof}[Proof of Lemma \ref{lem:TS_term3}]
	Let $\mathcal{F}_t$ represents the history till time $t$ that is,  $\mathcal{F}_{t}=(I_1,Y_{I_1},I_2,Y_{I_2}, \cdots I_t, Y_{I_t})$ and define $\mathcal{F}_{0}=\{\}$. Note that $\hat{p}_i(t)$, distribution of $\theta_{i}(t)$, and  either $E_{i}^{p}(t)$ is true or not is determined by  $\mathcal{F}_{t-1}$. Let $F_{t-1}$ be the instantiation of $\mathcal{F}_{t-1}$ where $\bar{E}_{i}^{p}$ is true. We define $q_{k,t} \coloneqq \Prob(\theta_{k}(t)<p_k+\epsilon|\mathcal{F}_{t-1}=F_{t-1})$ and  $\pmb{\theta}_{-k}(t)$ represents the vector $\pmb{\theta}(t)$ without $\theta_{k}(t)$. Let $\Theta_{i,k}(t)$ represents the collection of all possible values of $\pmb{\theta}(t)$ for which $A_{i,k}(t)$ and $\bar{E}_{i,k}^\theta (t)$ holds. Let $\Theta_{i,-k}(t)\coloneqq \{\pmb{\theta}_{-k}(t):\pmb{\theta}(t)\in \Theta_{i,k}(t)\}$. Let $M_i=\{j: l^{-1}_t(j)>l^{-1}_t(i)\}$ represents the arms after arm $i$ in cascade. Then,
	\begin{align*}
		\E&\left[\Indc\{A_{i,k}(t),\bar{E}^{p}_{i,k}(t),\bar{E}^{\theta}_{i,k}(t)\}\right]\\
		=&\E \left[ \Indc\{A_{i,k}(t), \bar{E}^{p}_{i,k}(t), \bar{E}^{\theta}_{i,k}(t)\} | \mathcal{F}_{t-1}=F_{t-1}\right]\\
		\le&  \Prob(\theta_j(t)\ge p_k+\epsilon ,\forall j\in M_i, I_t=i\\ &\qquad \pmb{\theta}_{-k}(t)\in \Theta_{i,-k}(t)| \mathcal{F}_{t-1}=F_{t-1})\\
		=& \Prob(\theta_k(t)\ge p_k+\epsilon| \mathcal{F}_{t-1}=F_{t-1}).\\
		&\E\left[\prod_{j<l^{-1}_t(i)}(1-X_{l_j}(t))|\pmb{\theta}_t\right].\\
		&\Prob(\theta_j(t)\ge p_k+\epsilon \forall j\in M_i/\{k\}, \\
		&\qquad \pmb{\theta}_{-k}(t)\in \Theta_{i,-k}(t)| \mathcal{F}_{t-1}=F_{t-1})\\
		=& (1-q_{k,t}).\E\left[\prod_{j<l^{-1}_i(t)}(1-X_{l_j}(t))|\pmb{\theta}_{-k}(t)\right].\\
		&\Prob(\theta_j(t)\ge p_k+\epsilon \forall j\in M_i/\{k\},\\ &\qquad \pmb{\theta}_{-k}(t)\in \Theta_{i,-k}(t)| \mathcal{F}_{t-1}=F_{t-1}) \numberthis \label{eq:TS_prt3_1}.
	\end{align*}
	Consider the instance where $\theta_{k}(t)$ is modified such that $\theta_{k}(t)<\theta_{i}(t)$ and $\pmb{\theta}_{-k}(t)$ is not modified, then 
	\begin{align*}
		\E\Big[\Indc\{&I_t=k,l^{-1}_t(k)<l^{-1}_t(i),\bar{E}^{\theta}_{i,k}(t),\\ 
		&\qquad \pmb{\theta}_{-k}(t)\in \Theta_{i,-k}(t)|\mathcal{F}_{t-1}=F_{t-1}\}\Big]\\
		>&\Prob(\theta_k(t)< p_k+\epsilon\le \theta_j(t), \forall j\in M_i/\{k\},I_t=k,\\
		&\qquad \pmb{\theta}_{-k}(t)\in \Theta_{i,-k}(t)| \mathcal{F}_{t-1}=F_{t-1}).\\
		=& \Prob(\theta_k(t)< p_k+\epsilon| \mathcal{F}_{t-1}=F_{t-1}).\\ &\E\left[\prod_{j<l^{-1}_k(t)}(1-X_{l_j})|\pmb{\theta}_{-k}(t)\right].\\
		&\Prob(\theta_j(t)\ge p_k+\epsilon ,\forall j\in M_i/\{k\},\\
		&\qquad \pmb{\theta}_{-k}(t)\in \Theta_{i,-k}(t)| \mathcal{F}_{t-1}=F_{t-1})\\
		\ge& q_{k,t}.\E\left[\prod_{j<l^{-1}_i(t)-1}(1-X_{l_j})|\pmb{\theta}_{-k}(t)\right].\\
		&\Prob(\theta_j(t)\ge p_k+\epsilon, \forall j\in M_i/\{k\},\\
		&\qquad \pmb{\theta}_{-k}(t)\in \Theta_{i,-k}(t)| \mathcal{F}_{t-1}=F_{t-1}) \numberthis \label{eq:TS_prt3_2}.
	\end{align*}
	From \eqref{eq:TS_prt3_1}, \eqref{eq:TS_prt3_2} we get
	
	\begin{align*}
		\E\Bigg[\sum_{t=1}^T\Indc&\{A_{i,k}(t),\bar{E}^{p}_{i,k}(t),\bar{E}^{\theta}_{i,k}(t)\}\Bigg]\\
		\le \sum_{t=1}^T& \E\Bigg[\frac{1-q_{k,t}}{q_{k,t}}\Indc\{I_t=k,l^{-1}_t(k)<l^{-1}_t(i),\\ &\bar{E}^{\theta}_{i,k}(t),\pmb{\theta}_{-k}(t)\in \Theta_{i,-k}(t)|\mathcal{F}_{t-1}=F_{t-1}\}\Bigg].
	\end{align*}
	
	Let $\tau_{k,s}$ be the time slot in which arm $k$ is chosen for s-th time, then we have
	\begin{align*}
		\sum_{t=1}^T\E&\Bigg[\frac{1-q_{k,t}}{q_{k,t}}\Indc\{ I_t=k,l^{-1}_t(k)<l^{-1}_t(i),\\
		&\bar{E}^{\theta}_{i,k}(t),\pmb{\theta}_{-k}(t)\in \Theta_{i,-k}(t)|\mathcal{F}_{t-1}=F_{t-1}\}\Bigg]\\
		&\quad \le \sum_{s=1}^{T} \E\left[\frac{1-q_{k,\tau_{k,s}}}{q_{k,\tau_{k,s}}} \right].
	\end{align*}
	
	\begin{lemma*}[Implied by Lemma 2.9 \cite{agrawal2017near}]
		If $\tau_{k,s}$ denote the time step at which s-th sample of arm $k$ is observed then we have 
		\begin{align*}
			\E&\left[\frac{1-q_{k,\tau_{k,s}}}{q_{k,\tau_{k,s}}} \right] \\
			\le &\begin{cases}
				\frac{3}{\epsilon} &\text{for } s<\frac{8}{\epsilon}\\
				\Theta\left(e^{-\epsilon^2s/2}+\frac{1}{(s+1)\epsilon^2}e^{-sD_{k}} +\frac{1}{e^{\epsilon^2s/4}-1}\right)  & \text{else, }
			\end{cases}
		\end{align*}
		where $D_k=KL(p_k,p_k+\epsilon)$.
	\end{lemma*}
	Now, we follow a similar analysis from Lemma 3.3 of \cite{zhong2021thompson} and improve the bound stated in \cite{zhong2021thompson}.
	
	\begin{align*}
		\E&\left[\frac{1-q_{k,\tau_{k,s}}}{q_{k,\tau_{k,s}}} \right]\\
		\le& \sum_{0\le s\le 8/\epsilon}\frac{3}{\epsilon}+ c_1.\sum_{8/\epsilon\le s\le T-1}e^{-\epsilon^2s/2}+\frac{1}{(s+1)\epsilon^2}e^{-sD_{k}}\\ &+\frac{1}{e^{\epsilon^2s/4}-1}\\
		\overset{(a)}{\le}& \frac{24}{\epsilon^2}+c_1.\sum_{s=1}^{\infty}e^{-\epsilon^2s/2} + c_1. \int_{8/\epsilon}^{T-1}\frac{1}{(s+1)\epsilon^2}e^{-2\epsilon^2s} ds\\ &+c_1\int_{8/\epsilon}^{T-1}\frac{1}{e^{\epsilon^2s/4}-1} ds\\
		\overset{(b)}{\le} &\frac{24}{\epsilon^2}+ \frac{2c_1}{\epsilon^2} +\frac{c_1e^{\epsilon^2/2}}{\epsilon^2}\left( \frac{1}{16\epsilon}\Indc\{16\epsilon<1\}+\frac{1}{e}\right)\\
		&+c_1\int_{8/\epsilon}^{T-1}\frac{1}{e^{\epsilon^2s/4}-1} ds,
	\end{align*}
	where $c_1$ is a constant. $(a)$ follows from the fact $KL(p,q)\ge \frac{|p-q|^2}{2}$. $(b)$ is obtained using the result from \cite{zhong2021thompson} and fact that $\sum_{t=1}^\infty e^{-at}\le \frac{1}{a}, a>0$.
	Consider 	
	\begin{align*}
		\int_{8/\epsilon}^{T-1}\frac{1}{e^{\epsilon^2s/4}-1} ds\le& \int_{8/\epsilon}^{\infty}\frac{1}{e^{\epsilon^2s/4}-1} ds\\
		=&\frac{8}{\epsilon}-\frac{4\log (e^{2\epsilon}-1)}{\epsilon^2}.
	\end{align*}
	Therefore,
	\begin{align*}
		\E&\left[\sum_{t=1}^T\Indc\{A_{i,k}(t),\bar{E}^{p}_{i,k}(t),\bar{E}^{\theta}_{i,k}(t)\}\right]\\
		\le& \frac{24}{\epsilon^2}+ \frac{2c_1}{\epsilon^2} +\frac{c_1e^{\epsilon^2/2}}{\epsilon^2}\left( \frac{1}{16\epsilon}\Indc\{16\epsilon<1\}+\frac{1}{e}\right)\\ &+ \frac{8c_1}{\epsilon}-\frac{4c_1\log (e^{2\epsilon}-1)}{\epsilon^2}.
	\end{align*}
\end{proof}

\end{document}